\documentclass[letterpaper]{article} % DO NOT CHANGE THIS
\usepackage{aaai25}  % DO NOT CHANGE THIS
\usepackage{times}  % DO NOT CHANGE THIS
\usepackage{helvet}  % DO NOT CHANGE THIS
\usepackage{courier}  % DO NOT CHANGE THIS
\usepackage[hyphens]{url}  % DO NOT CHANGE THIS
\usepackage{graphicx} % DO NOT CHANGE THIS
\usepackage{natbib}  % DO NOT CHANGE THIS AND DO NOT ADD ANY OPTIONS TO IT
\usepackage{caption} % DO NOT CHANGE THIS AND DO NOT ADD ANY OPTIONS TO IT
\usepackage{algorithm}
\usepackage{algorithmic}
\usepackage{amsmath}
\usepackage{colortbl}
\usepackage{amssymb}
\usepackage{newfloat}
\usepackage{listings}
\DeclareCaptionStyle{ruled}{labelfont=normalfont,labelsep=colon,strut=off} % DO NOT CHANGE THIS
\floatstyle{ruled}
\newfloat{listing}{tb}{lst}{}
\floatname{listing}{Listing}
\title{Optimizing Human Pose Estimation Through Focused Human and Joint Regions}
\author{
    Yingying Jiao\textsuperscript{\rm 1,2},
    Zhigang Wang\textsuperscript{\rm 3}\thanks{Corresponding authors.},
    Zhenguang Liu\textsuperscript{\rm 4,5}\footnotemark[1],
    Shaojing Fan\textsuperscript{\rm 6},\\
    Sifan Wu\textsuperscript{\rm 1,2}\footnotemark[1],
    Zheqi Wu\textsuperscript{\rm 3},
    Zhuoyue Xu\textsuperscript{\rm 3},
}
\affiliations{
    \textsuperscript{\rm 1}College of Computer Science and Technology, Jilin University\\
    \textsuperscript{\rm 2}Key Laboratory of Symbolic Computation and Knowledge Engineering of Ministry of Education, Jilin University\\
    \textsuperscript{\rm 3}College of Computer Science and Technology, Zhejiang Gongshang University\\
    \textsuperscript{\rm 4}The State Key Laboratory of Blockchain and Data Security, Zhejiang University\\
    \textsuperscript{\rm 5}Hangzhou High-Tech Zone (Binjiang) Institute of Blockchain and Data Security\\
    \textsuperscript{\rm 6}School of Computing, National University of Singapore\\
    jiaoyy21@mails.jlu.edu.cn, wangzhigang2024@gmail.com, liuzhenguang2008@gmail.com, fanshaojing@gmail.com,\\ wusifan2021@gmail.com, chasewoo17@gmail.com, 1069516849xzyy@gmail.com
    
}

\begin{document}

\definecolor{mygray}{gray}{.9}

\maketitle

\begin{abstract}
Human pose estimation has given rise to a broad spectrum of novel and compelling applications, including \textit{action recognition}, \textit{sports analysis}, as well as \textit{surveillance}. However, accurate video pose estimation remains an open challenge. One aspect that has been overlooked so far is that existing methods learn motion clues from all pixels rather than focusing on the target human body, making them easily misled and disrupted by unimportant information such as \textit{background changes} or \textit{movements of other people}. Additionally, while the current Transformer-based pose estimation methods has demonstrated impressive performance with global modeling, they struggle with local context perception and precise positional identification. 

In this paper, we try to tackle these challenges from three aspects: (1) We propose a bilayer Human-Keypoint Mask module that performs coarse-to-fine visual token refinement, which gradually zooms in on the target human body and keypoints while masking out unimportant figure regions. (2) We further introduce a novel deformable cross attention mechanism and a bidirectional separation strategy to adaptively aggregate spatial and temporal motion clues from constrained surrounding contexts. (3) We mathematically formulate the deformable cross attention, constraining that the model focuses solely on the regions centered at the target person body. Empirically, our method achieves state-of-the-art performance on three large-scale benchmark datasets. A remarkable highlight is that our method achieves an 84.8 mean Average Precision (mAP) on the challenging \textit{wrist} joint, which significantly outperforms the 81.5 mAP achieved by the current state-of-the-art method on the PoseTrack2017 dataset.  
\end{abstract}


\section{Introduction}

Human pose estimation, as a fundamental problem in the realm of computer vision and artificial intelligence~\cite{wang2022contextual, geng2023human}, involves accurately identifying the anatomical keypoints of human bodies. Precise pose estimation is the key for the success of a machine as it paves the way for machines to accurately interpret human movements and behaviors. Accordingly, human pose estimation spans a wide range of applications from \textit{action recognization}, \textit{movement tracking}, to \textit{augmented reality}~\cite{yang2023action, tse2019no, su2021motion, wu2024pose, liu2022copy}.

A plethora of research has been dedicated to the field of pose estimation on still images, evolving from early methods employing tree-based and random forest models~\cite{wang2008multiple,sapp2010cascaded} to current methodologies utilizing convolutional neural networks~\cite{sun2019hrnet} and Transformers~\cite{li2021tokenpose}. Despite their excellent performance on still images, applying these methods directly to video pose estimation leads to significant performance degradation due to the exclusive characteristics in videos, such as \textit{rapid movement} and \textit{video defocus}, which are frequently encountered in videos but absent in static images~\cite{zhao2018recognize}.

\begin{figure*}[t]
\centering
\includegraphics[width=.93\linewidth]{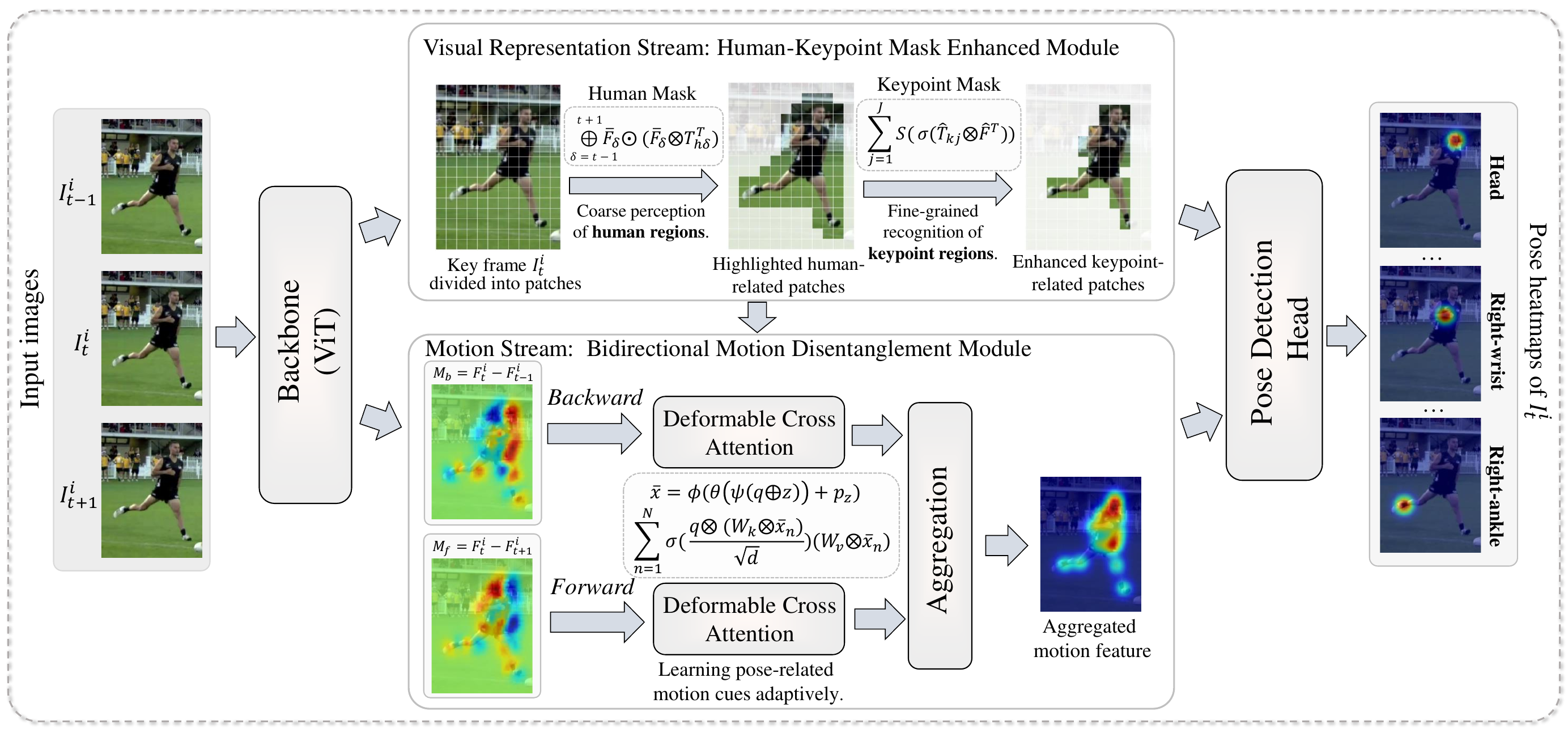}
\caption{A high-level overview of our proposed VREMD, which utilizes a dual-stream architecture to collaboratively process and integrate complementary visual and motion features. The visual representation stream executes progressive enhancement of human keypoint-related features to achieve precise location recognition. The motion stream performs adaptive pose-related motion disentanglement through the novel deformable cross attention. $\{\boldsymbol{F}_{t-1}^{i}, \boldsymbol{F}_{t}^{i}, \boldsymbol{F}_{t+1}^{i}\}$ denote the visual features of three input frames \(\{I_{t-1}^{i}, I_{t}^{i}, I_{t+1}^{i}\}\) output by backbone network.} 
\label{fig:head}
\end{figure*}

To address this issue, substantial studies have emerged that leverage temporal continuity to extract rich semantic visual contexts for human pose estimation in videos. Current methods can be roughly categorized into two main branches. One line of research~\cite{bertasius2019posewarper,liu2021dcpose} aggregates temporal information from neighboring frames for video pose estimation, employing CNN-based architectures and pose calibration. Fueled by the development of Transformers~\cite{dosovitskiy2020imagevit,vaswani2017transformers}, another line of studies~\cite{jin2022otpose, he2024dsta} strive to integrate attention mechanisms into model construction, yielding impressive results and showcasing their immense potential. However, a limitation inherent in existing Transformer-based methods~\cite{jin2022otpose} lies in their inability to effectively manage local dependencies. This limitation poses a notable challenge for visual perception tasks such as pose estimation, which require precise local positioning.

Following thorough experimentation and empirical investigation, we uncover two insights: (1) Existing methods~\cite{liu2022fami, feng2023tdmi, wu2024joint} struggle to handle subtle pose changes, particularly in challenging scenarios with occlusions or motion blur. This may stem from the fact that current methods tend to capture temporal dynamics pixel-by-pixel rather than focusing solely on target human regions, leading to them being distracted by unuseful cues such as background changes or pixels far from the target person. (2) Additionally, previous studies~\cite{liu2021dcpose,liu2022fami} adopting multiple sets of fixed deformable convolutions with varying dilation rates, which neglect the importance of adaptive scale selection.

Inspired by these, we propose a dual-stream framework, which executes \textbf{\underline{V}}isual \textbf{\underline{R}}epresentation \textbf{\underline{E}}nhancement and \textbf{\underline{M}}otion \textbf{\underline{D}}isentanglement (VREMD) for human pose estimation in videos. Technically, we embrace three novel designs to tackle the challenge. (1) We propose a two-step human-keypoint mask module for coarse-to-fine visual enhancement, which progressively refines extracted representations from the human body and keypoints perspectives. (2) We further introduce a bidirectional decoupled module tailored for adaptively disentangling motion cues of the target person from unnecessary visual elements. (3) Furthermore, we mathematically formulate a deformable cross attention mechanism that constrains the model to focus exclusively on regions circumscribing the target human body.

\begin{figure*}[t]
\centering
\includegraphics[width=\linewidth,keepaspectratio]{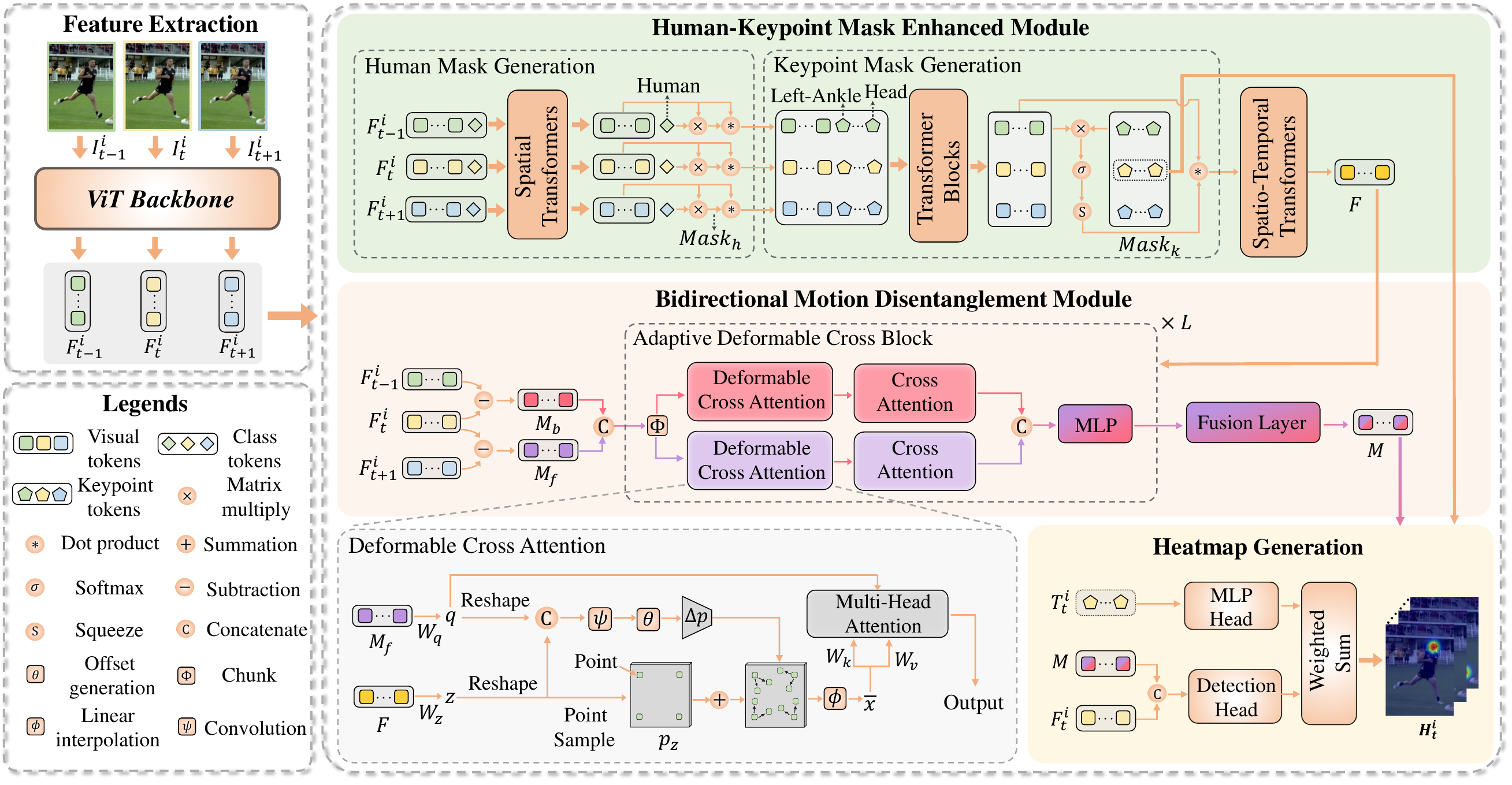}
\caption{The overall pipeline of our VREMD framework. Given an input sequence $\{ I_{t-1}^{i},  I_{t}^{i} , I_{t+1}^{i}\}$, our goal is to estimate the human pose of the key frame $I_{t}^{i}$. We initially extract the visual features via a ViT backbone, and then feed them into the Human-Keypoint Enhanced module and the Bidirectional Motion Disentanglement module to obtain $\boldsymbol{T}_{t}^{i}$ and $\boldsymbol{M}$. Finally, the outputs derived from different heads are combined through a weighted sum to arrive at the final predicted pose heatmap $\boldsymbol{H}_{t}^{i}$.} 
\label{fig:framework}
\end{figure*}

Our framework exemplifies the collaborative advantage between local spatial focus and adaptive temporal clues extraction, opening up possibilities for rethinking the pose estimation task from emphasizing on the target human body and masking out the irrelevant spatio-temporal contexts. To evaluate the efficacy of our method, we conduct extensive experiments on three public benchmarks, achieving state-of-the-art performance. The key contributions of our method are summarized as follows:
\begin{itemize}
    \item We present a dual-stream framework that integrates visual enhancement and motion disentanglement to highlight target human areas and fliter other non-essential regions for human pose estimation.
    \item We creatively introduce a deformable cross attention to disentangle pose-related motion cues, harnessing bidirectional temporal dynamics and enabling the model to robustly handle complex pose variations of the target human. 
    \item Empirically, our method achieves state-of-the-art performance on three large-scale benchmarks, and overall provides insights into integrating Transformer-based methods with region-specific enhancement strategies to boost their local localization capabilities. 
\end{itemize}

\section{Our Method}

\textbf{Preliminaries.} Our method follows the top-down paradigm, which first extracts each individual person from an image and then estimates their poses. Specifically, we first utilize an object detector to extract the bounding box for person \(i\) in a video frame \(I_{t}\) that is to be detected. Subsequently, we expand the bounding box by 25$\%$ and crop the same person in the adjacent frames (\textit{i.e.}, \(I_{t-1}\) and \(I_{t+1}\)). As a result, we obtain a sequence of consecutive frames for person \(i\): 
\(\boldsymbol{\mathcal{I}}_{\boldsymbol{t}}^{\boldsymbol{i}} = \{I_{t-1}^{i}, I_{t}^{i}, I_{t+1}^{i}\}\). Given a sequence of video frames $\boldsymbol{\mathcal{I}}_{\boldsymbol{t}}^{\boldsymbol{i}}$ that includes the key frame $I_{t}^{i}$ and the auxiliary frames $I_{t-1}^{i}$ and $I_{t+1}^{i}$, our target is to detect the human pose within $I_{t}^{i}$. We aim to strengthen the utilization of supplementary temporal information in auxiliary frames by employing incremental visual representation enhancement and adaptively disentangling useful motion information, thus tackling the common issue of existing methods being interfered with by irrelevant information regarding the target human.

\textbf{Method overview.} The overview pipeline of our proposed VREMD is depicted in Figure~\ref{fig:framework}. VREMD constructs a dual-stream architecture with inter-module communication that enhances both visual features and captures meaningful motion cues. Specifically, VREMD incorporates two distinct modules: a Human-Keypoint Mask Enhanced module (HKME) and a Bidirectional Motion Disentanglement module (BMD). 
First, we utilize a Vision Transformer backbone to extract visual features $\{\boldsymbol{F}_{t-1}^{i}, \boldsymbol{F}_{t}^{i}, \boldsymbol{F}_{t+1}^{i}\}$ from the input frame sequence $\boldsymbol{\mathcal{I}}_{\boldsymbol{t}}^{\boldsymbol{i}}$, which are then simultaneously fed into both the HKME and BMD modules. The HKME generates dual masks for a coarse-to-fine representation refinement, resulting in enhanced feature $\boldsymbol{F}$ and key frame keypoint tokens $\boldsymbol{T}_{t}^{i}$. The BMD computes the motion features and, utilizing $\boldsymbol{F}$ as a constraint, dynamically derives joint-related motion contexts to produce the filtered $\boldsymbol{M}$. Finally, the keypoint heatmaps $\boldsymbol{H}_{k}$ from key frame tokens $\boldsymbol{T}_{t}^{i}$ via an MLP and the heatmaps $\boldsymbol{H}_{m}$ decoded from $\boldsymbol{M}$ and key frame features $\boldsymbol{F}_{t}^{i}$ are weighted, summed, and combined to produce the final pose estimation $\boldsymbol{H}_{t}^{i}$. The following sections will elaborate on the two key components in detail.

\subsection{Human-Keypoint Mask Enhanced Module}

Despite the Transformers architecture achieving remarkable success in various fields~\cite{dosovitskiy2020imagevit, vaswani2017transformers}, its application in video pose estimation has been limited. Given the significant potential demonstrated by this architecture in other visual perception tasks~\cite{zheng2021rethinking,li2022exploring}, we seek to design a novel Transformer-based framework specially tailored for video pose detection. A naive approach to aggregate unique temporal cues from a video would be to concatenate features across multiple frames for full-token computation. Yet, such a straightforward treatment strategy faces two issues: excessive capture of redundant information between adjacent frames, and a lack of focus on task-relevant tokens. 

Inspired by previous work~\cite{strudel2021segmenter,li2021tokenpose}, we propose a Human-Keypoint Mask Enhanced module with a progressive refinement architecture, addressing the aforementioned issues through three steps: 
(1) We generate a human mask to coarsely enhance the perception of the target human. (2) We produce a keypoint mask to achieve finer filtering of keypoint-related features. (3) We utilize spatio-temporal networks to aggregate the highlighted spatio-temporal cues of these visual features. 
This step-by-step optimization strategy can discern articular visual tokens, simulating the capability of localized identification, which promotes precise pose estimation.

\textbf{Human mask.} Given a visual feature sequence $\{\boldsymbol{F}_{t-1}^{i}, \boldsymbol{F}_{t}^{i}, \boldsymbol{F}_{t+1}^{i}\} \in \mathbb{R}^{3 \times N \times D }$ output by the ViT backbone, we concatenate a learnable class token $\boldsymbol{T}_{h} \in \mathbb{R}^{3 \times 1 \times D }$ with a category of human to each feature. These features then individually pass through cascaded Transformer blocks for intra-frame spatial similarity computation. We separate the result into human token $\boldsymbol{T}_{h}$ and visual features $\overline{\boldsymbol{F}} \in \mathbb{R}^{3 \times N \times D }$. After transposing the human token, we perform matrix multiplication to obtain the human mask $\boldsymbol{Mask}_{h} \in \mathbb{R}^{3 \times N \times 1 }$. Finally, we secure a coarsely selected feature $\boldsymbol{F}_{c} \in \mathbb{R}^{3 \times N \times D }$ by executing element-wise dot product between the human mask $\boldsymbol{Mask}_{h}$ and the visual feature $\overline{\boldsymbol{F}}$, utilizing broadcasting. The above operations can be formulated as:
\begin{equation} 
  \begin{aligned}
        \boldsymbol{F}_{c}=\bigoplus_{\delta = t-1 }^{t+1}  \overline{\boldsymbol{F}}_{\delta} \odot \underbrace{( \overline{\boldsymbol{F}}_{\delta} \otimes \boldsymbol{T}_{h\delta}^{T} )}_{\boldsymbol{Mask}_{h\delta}} ,
  \end{aligned}
\end{equation}
where $\bigoplus$, $\delta$, $\odot$, $\otimes$, and $\boldsymbol{T}^{T}$ denote concatenation, temporal index of frames, dot product, matrix multiplication, the transpose of $\boldsymbol{T}$, respectively.

\textbf{Keypoint mask.} In pursuit of more precise keypoint-related feature enhancement, we employ additional auxiliary tokens to accurately localize spatial positions by integrating multi-frame representations in the spatio-temporal domain. We concatenate the learnable keypoint tokens $\boldsymbol{T}_{k} \in \mathbb{R}^{3 \times J \times D }$ (Note that $J$ is the number of keypoints) to the coarsely selected feature $\boldsymbol{F}_{c}$ and separate the multi-frame features, which are then linked along the token dimension and fed into Transformer blocks for spatio-temporal learning. Subsequently, we split the visual features and keypoint tokens from the output and gather them over multiple frames, resulting in multi-frame features $\widehat{\boldsymbol{F}} \in \mathbb{R}^{(3 \cdot N) \times D }$ and multi-frame keypoint tokens $\widehat{\boldsymbol{T}}_{k} \in \mathbb{R}^{(3 \cdot J) \times D }$. After transposing the multi-frame features, we perform matrix multiplication with the multi-frame keypoint tokens to produce the keypoint confidence map $\boldsymbol{Map} \in \mathbb{R}^{(3 \cdot J) \times (3 \cdot N) }$. We apply the softmax function to compute element-wise weights for the map $\boldsymbol{Map}$, and summing along the second-to-last dimension followed by transposition yields the keypoint mask $\boldsymbol{Mask}_{k} \in \mathbb{R}^{(3 \cdot N) \times 1 }$:
\begin{equation} 
  \begin{aligned}
        \boldsymbol{Mask}_{k} = \sum_{{j =1}}^{J}  S(\sigma (\widehat{\boldsymbol{T}}_{kj} \otimes \widehat{\boldsymbol{F}}^{T}  )),
  \end{aligned}
\end{equation}
where $j \in \{1, \dots, J\}$, $S(\cdot)$ , \(\sigma(\cdot)\), and $\otimes$ denote the keypoint index, squeeze operation, softmax function, and matrix multiplication, respectively. The keypoint mask is element-wise multiplied with the multi-frame features $\widehat{\boldsymbol{F}}$ to create the refined filtered features $\boldsymbol{F}_{f} \in \mathbb{R}^{(3 \cdot N) \times D }$.

\textbf{Spatio-temporal aggregation.} To fully leverage the refined representation information, we perform decoupled spatio-temporal feature aggregation through the spatio-temporal Transformers. Specifically, we first separate the refined filtered features $\boldsymbol{F}_{f}$ and undertake frame-level spatial modulation. Then, each token is concatenated with its corresponding token in the temporal domain to undergo temporal modulation, resulting in $\overline{\boldsymbol{F}}_{f} \in \mathbb{R}^{(3 \cdot N) \times D }$. Finally, we adopt an MLP to execute token dimensionality reduction on $\overline{\boldsymbol{F}}_{f}$ to attain spatio-temporal aggregation of multi-frame features, leading to the enhanced feature $\boldsymbol{F} \in \mathbb{R}^{N \times D}$.

\subsection{Bidirectional Motion Disentanglement Module}

To extract useful complementary information from auxiliary frames, prior methods~\cite{liu2021dcpose,feng2023tdmi} implicitly model feature residuals to capture motion evidence. The common practice among these paradigms is to directly concatenate the computed multiple motion features for convolution after their calculation, which considers temporal continuity but overlooks insights from the temporal direction. We observe that, from the perspective centered around the key frame, the essential temporal details that need to be focused on actually originate from two different directions, namely forward and backward. Considering this intrinsic factor, we design a bidirectional separation strategy to decouple the continuous motion into parallel forward and backward motion trajectories. Furthermore, existing methods do not differentiate motion clues in the spatial dimension, which can lead to learning pose-irrelevant information (\textit{e.g.}, background, other people, etc.) that can disrupt detection. Moreover, existing methods heavily rely on deformable convolutions for local motion calibration, potentially leading to models that are overly tailored and limiting their compatibility with Transformer-based architectures. To tackle these challenges, we introduce deformable cross attention (DCA) for the first time and create the Adaptive Deformable Cross block by employing it, which adaptively captures pose-related motion dynamics.

\begin{table*}[t] 
    \centering
    \fontsize{9pt}{9pt}\selectfont
    \resizebox{0.95\linewidth}{!}{
    \begin{tabular}{l|ccccccc|c}
         \hline
         Method & Head & Shoulder & Elbow & Wrist & Hip & Knee & Ankle & Mean \\
         \hline
         PoseTracker~\cite{girdhar2018posetracker} & 67.5& 70.2& 62.0& 51.7& 60.7& 58.7& 49.8 & 60.6 \\ 
          PoseFlow~\cite{xiu2018pose} & 66.7 & 73.3 & 68.3 & 61.1 & 67.5 & 67.0 & 61.3 & 66.5 \\
          JointFlow~\cite{doering2018jointflow}& - &- &-& - &-& -& - &69.3\\
          FastPose~\cite{zhang2019fastpose}& 80.0 &80.3 &69.5& 59.1 &71.4 &67.5 &59.4 &70.3  \\
          TML++~\cite{hwang2019tml} & - & - & - &-&-&-& - & 71.5 \\
          Simple (R-50)~\cite{xiao2018simplebaseline}& 79.1 &80.5& 75.5& 66.0 &70.8& 70.0 &61.7& 72.4 \\
          Simple (R-152)~\cite{xiao2018simplebaseline} & 81.7 &83.4 &80.0 &72.4 &75.3 &74.8 &67.1 &76.7 \\
          STEmbedding~\cite{jin2019stembedding} & 83.8 & 81.6 & 77.1 &70.0 &77.4 &74.5 &70.8 &77.0 \\
                             HRNet~\cite{sun2019hrnet} & 82.1 & 83.6 & 80.4 & 73.3 & 75.5 & 75.3 & 68.5 & 77.3 \\
                             MDPN~\cite{guo2018mdpn}  & 85.2 & 88.5 & 83.9 & 77.5 & 79.0 & 77.0 & 71.4 & 80.7 \\
                             CorrTrack~\cite{rafi2020self} &  86.1 & 87.0 &  83.4  & 76.4  & 77.3  & 79.2 &  73.3  & 80.8  \\
                             Dynamic-GNN~\cite{yang2021dynamicgnn}& 88.4& 88.4& 82.0 &74.5& 79.1& 78.3 &73.1& 81.1 \\
                             PoseWarper~\cite{bertasius2019posewarper}  &  81.4 &88.3 &83.9& 78.0& 82.4 &80.5 &73.6 & 81.2  \\
                             DCPose~\cite{liu2021dcpose} & 88.0 & 88.7 & 84.1 & 78.4 &  83.0 & 81.4 & 74.2 & 82.8  \\
                             DetTrack~\cite{wang2020dettrack}& 89.4& 89.7& 85.5 &79.5 &82.4 &80.8 &76.4 &83.8 \\
                             SLT-Pose~\cite{gai2023sltpose} & 88.9& 89.7 &85.6 &79.5 &84.2 &83.1 & 75.8& 84.2\\
                             HANet~\cite{jin2023hanet} & 90.0 &90.0 &85.0 & 78.8 & 83.1 & 82.1 & 77.1&84.2\\
                             KPM~\cite{fu2023kpm} & 89.5 & 90.0 & 87.6 & 81.8  & 81.1 & 82.6 & 76.1 &84.6\\
                             M-HANet~\cite{jin2024mhanet} & 90.3 & 90.7& 85.3& 79.2& 83.4& 82.6& 77.8 &84.8\\
                             FAMI-Pose~\cite{liu2022fami} & 89.6  & 90.1 & 86.3 & 80.0 & 84.6 & 83.4 & 77.0 & 84.8   \\
                             DSTA~\cite{he2024dsta} & 89.3  & 90.6 & 87.3 & 82.6 & 84.5 & 85.1 & 77.8 & 85.6   \\
                             TDMI-ST~\cite{feng2023tdmi}  & \textbf{90.6} & 91.0 & 87.2 & 81.5 &  85.2 & 84.5 & 78.7 & 85.9 \\
                            \hline
                              \rowcolor{mygray} \textbf{VREMD (Ours)} & 89.9 & \textbf{91.4} & \textbf{88.8} & \textbf{84.8} & \textbf{88.5} & \textbf{87.8} &  \textbf{81.0} & \textbf{87.6} \\
                 
        \hline

    \end{tabular}
    }
    \caption{Comparisons with the state-of-the-art methods for video pose estimation on the validation sets of the \textbf{PoseTrack2017}~\cite{iqbal2017posetrack} dataset. Note that we aggregate temporal information from neighboring frames (\textit{i.e.}, one frame to the left and one to the right).} 
    \label{table:compare_17}
\end{table*}

\textbf{Adaptive Deformable Cross block.} Given the features $\{\boldsymbol{F}_{t-1}^{i}, \boldsymbol{F}_{t}^{i}, \boldsymbol{F}_{t+1}^{i}\}$ from the backbone, we subtract $\boldsymbol{F}_{t}^{i}$ from both $\boldsymbol{F}_{t+1}^{i}$ and $\boldsymbol{F}_{t-1}^{i}$ to obtain $\{\boldsymbol{M}_{f}, \boldsymbol{M}_{b}\}$. Adaptive Deformable Cross blocks (ADC) take the concatenation of $\boldsymbol{M}_{f}$ and $\boldsymbol{M}_{b}$, along with the enhanced feature $\boldsymbol{F}$ from HKME. After entering the ADC block, $\boldsymbol{M}_{f}$ and $\boldsymbol{M}_{b}$ are first split, and then pass through a dual-branch structure that includes a deformable cross attention (DCA) and a cross attention. The results from the dual branches are concatenated and sent into an MLP for nonlinear transformation. After the final block, a fusion layer is applied to integrate the bidirectional motion features to obtain an aggregated motion representation $\boldsymbol{M}$.

\textbf{Deformable cross attention.} Our deformable cross attention (DCA) predicts multiple offsets at a single point, rather than predicting offsets at each point of the kernel as in the case of deformable convolution. This endows it with a stronger ability to characterize the relationships between elements and to flexibly handle different scales. The concept of our cross mechanism is realized by incorporating the enhanced feature $\boldsymbol{F}$ as a constraint to control the generation of offsets in the spatial domain, ensuring that only a subset of motion features are selected as keys and values for attention computation. Specifically, the DCA can be represented by the following formulas:
\begin{equation} 
  \begin{aligned}
        q &= W_{q} \otimes x, &
        z &= W_{z} \otimes \boldsymbol{F},\\
        \bigtriangleup p &= \theta(\psi ( q \oplus z)),&
        \overline{x} &=   \phi (\bigtriangleup p + p_{z}),\\
  \end{aligned}
\end{equation}\label{dca_1}
\begin{equation} 
  \begin{aligned}
        \text{DCA}(x,z,p_{z})  &= \sum_{n =1 }^{N} \sigma (\frac{q\otimes (W_{k}\otimes\overline{x}_{n} )^{T} }{\sqrt{d} } )(W_{v}\otimes\overline{x}_{n} ) ,
  \end{aligned}
\end{equation}\label{dca_2} 
where $x$, $q$, $\bigtriangleup p$, $p_{z}$, $\overline{x}$, $N$, and $d$ are motion features $\boldsymbol{M}_{f}$ or $\boldsymbol{M}_{b}$, query, point offset, reference points from $z$, sample features, number of sampling points, and embedding dimension, respectively. $\otimes$, $\oplus$, $\psi(\cdot)$, $\theta(\cdot)$, $\phi(\cdot)$, and $\sigma(\cdot)$ denote the operations of matrix multiplication, concatenation, convolution, offset generation, bilinear interpolation, softmax, respectively. $W_{q}$, $W_{k}$, $W_{v}$, and $W_{z}$ are all learnable mapping matrices. The offset $\bigtriangleup p$ generated under the constraint of $\boldsymbol{F}$, ensures the filtering of spatial regions related to the human joints within the global domain, thereby facilitating adaptive motion cue extraction from motion features.

\textbf{Heatmap generation.} We first split the key frame keypoint tokens \(\boldsymbol{T}_{t}^{i}\) from \(\widehat{\boldsymbol{T}}_{k}\) and then transform them into \(\boldsymbol{H}_{k}\) through an MLP and reshaping. By aggregating \(\boldsymbol{M}\) and \(\boldsymbol{F}_{t}^{i}\) and up-sampling, we obtain \(\boldsymbol{H}_{m}\). The final pose heatmaps \(\boldsymbol{H}_{t}^{i}\) are derived by adding \(\boldsymbol{H}_{k}\) and \(\boldsymbol{H}_{m}\) with equal weights.

\textbf{Loss function.} We adopt the established pose heatmap loss \(\mathcal{L}_{\mathrm{H}}\) to supervise the final predicted pose heatmaps \(\boldsymbol{H}_{t}^{i}\) to converge to the ground truth pose heatmaps \(\boldsymbol{G}_{t}^{i}\):
\begin{equation}\label{loss}
\begin{aligned}
 \mathcal{L}_{\mathrm{H}} = \left\| \boldsymbol{H}_{t}^{i} - \boldsymbol{G}_{t}^{i}\right\|_{2}^{2}.
\end{aligned}
\end{equation}

\section{Experiments}

\subsection{Experimental Settings}

\textbf{Datasets.} PoseTrack has become a crucial dataset in video-based human pose estimation benchmarks. \textbf{PoseTrack2017}\cite{iqbal2017posetrack} introduces 250 training videos and 50 validation videos, with 80,144 pose annotations across 15 key points. \textbf{PoseTrack2018}\cite{andriluka2018posetrack} expands to 593 training and 170 validation videos, totaling 153,615 annotations. \textbf{PoseTrack2021}~\cite{doering2022posetrack21} further enriches the dataset, particularly improving the representation of smaller figures and crowded scenes, reaching 177,164 pose annotations, with recalibrated joint visibility flags to better address occlusions.

\textbf{Evaluation metric.} To evaluate the efficacy of our proposed model in pose estimation, we calculate the average precision (AP) for each joint and then aggregate these values to obtain the mean average precision (mAP).

\textbf{Implementation details.} Our VREMD framework is realized utilizing PyTorch. For feature extraction on single frames, we adopt the most primitive Vision Transformer (ViT-L) architecture~\cite{dosovitskiy2020imagevit,xu2022vitpose}, pre-trained on the COCO dataset~\cite{lin2014microsoftcoco}, as our backbone. The input image size is fixed at 256\(\times\)192. We integrate a series of data augmentation techniques, consistent with methodologies employed in previous works~\cite{bertasius2019posewarper,liu2021dcpose}, comprising random rotation \([-45^\circ, 45^\circ]\), random scale [0.65, 1.35], truncation (half body), and flipping during training. The number of input frames is set to 3, consisting of one key frame accompanied by two auxiliary frames sourced from preceding and succeeding neighbors, respectively. This configuration mirrors that of DCPose~\cite{liu2021dcpose}, rather than employing the five frame input as seen in TDMI~\cite{feng2023tdmi} and FAMI-Pose~\cite{liu2022fami}. Our model is trained on a single RTX 4090 GPU for 20 epochs with the backbone frozen. We utilize the AdamW optimizer with an initial learning rate of 2e-3, which is then reduced by a factor of ten at the 16th epoch.

\begin{table*}[t] 
    \centering
    \fontsize{9pt}{9pt}\selectfont
    \resizebox{0.99\linewidth}{!}{
    \begin{tabular}{l|ccccccc|c}
        \hline
         Method & Head & Shoulder & Elbow & Wrist & Hip & Knee & Ankle & Mean \\
         \hline
        % STAF~\cite{raaj2019staf} & -& -& -& 64.7 &- &- &62.0 &70.4\\
        AlphaPose~\cite{fang2017alphapose} & 63.9 &78.7 &77.4 &71.0 &73.7 &73.0 &69.7 &71.9\\
        TML++~\cite{hwang2019tml} & - &-& -& - &- &-& - &74.6\\
        MDPN~\cite{guo2018mdpn}& 75.4& 81.2  &79.0 & 74.1  &72.4 & 73.0  &69.9 & 75.0\\
        PGPT~\cite{bao2020pgpt}& - &- &-& 72.3 &-& -& 72.2& 76.8\\
        Dynamic-GNN~\cite{yang2021dynamicgnn}&  80.6 & 84.5&  80.6&  74.4 & 75.0&  76.7 & 71.8 & 77.9 \\
        
        PoseWarper~\cite{bertasius2019posewarper}  & 79.9 & 86.3  &82.4  & 77.5 & 79.8  &78.8 & 73.2  &79.7 \\
                             PT-CPN++~\cite{yu2018multi}   & 82.4& 88.8 &86.2 &79.4& 72.0 &80.6& 76.2& 80.9 \\
                             DCPose~\cite{liu2021dcpose}  &84.0& 86.6& 82.7& 78.0& 80.4 &79.3 &  73.8& 80.9 \\
                             DetTrack~\cite{wang2020dettrack}& 84.9  &87.4  &84.8 & 79.2  &77.6  &79.7  &75.3 & 81.5 \\
                             FAMI-Pose~\cite{liu2022fami}  &85.5& 87.7 &84.2& 79.2& 81.4 &81.1 &  74.9& 82.2  \\
                             HANet~\cite{jin2023hanet} & 86.1 & 88.5& 84.1& 78.7&79.0 & 80.3& 77.4& 82.3 \\
                             M-HANet~\cite{jin2023hanet}& 86.7& 88.9& 84.6& 79.2& 79.7& 81.3& 78.7& 82.7\\
                             KPM~\cite{fu2023kpm} & 85.1& 88.9& \textbf{86.4}& 80.7& 80.9& 81.5& 77.0& 83.1\\
                             DSTA~\cite{he2024dsta} & 85.9  & 88.8 & 85.0 & 81.1 & 81.5 & 83.0 &  77.4 & 83.4   \\
                             TDMI-ST~\cite{feng2023tdmi}  &\textbf{86.7}& 88.9 &85.4 &80.6 &82.4& 82.1&  77.6& 83.6  \\       
                             \hline
                             \rowcolor{mygray} \textbf{VREMD (Ours)} & \textbf{86.7} & \textbf{89.3} & \textbf{85.6} & \textbf{82.1} & \textbf{85.0} & \textbf{83.9} &  \textbf{79.3} &  \textbf{84.6} \\
        \hline
    \end{tabular}
    }
    \caption{Comparisons with the state-of-the-art methods for video pose estimation on the validation sets of the \textbf{PoseTrack2018}~\cite{andriluka2018posetrack} dataset.} 
    \label{table:compare_18}
\end{table*}

\begin{table*}[t] 
    \centering
    \fontsize{9pt}{9pt}\selectfont
    \resizebox{0.99\linewidth}{!}{
    \begin{tabular}{l|ccccccc|c}
        \hline
         Method & Head & Shoulder & Elbow & Wrist & Hip & Knee & Ankle & Mean \\
        \hline
        Tracktor++ w. poses~\cite{bergmann2019tracking} &  - & - &  -  & -  & -  & - &  -  & 71.4\\
        CorrTrack~\cite{rafi2020self} &  - & - &  -  & -  & -  & - &  -  & 72.3  \\  
        Tracktor++ w. corr.~\cite{bergmann2019tracking}& - &- &- &- &- &- &- &73.6 \\
        DCPose~\cite{liu2021dcpose} &83.2 &84.7& 82.3 &78.1 &80.3 &79.2 & 73.5& 80.5  \\
                             FAMI-Pose~\cite{liu2022fami}  &83.3 &85.4 &82.9 &78.6 &81.3 &80.5 & 75.3 &81.2 \\
                             DSTA~\cite{he2024dsta} & \textbf{87.5}  & 87.0 & 84.2 & 81.4 & 82.3 & 82.5 &  77.7 & 83.5   \\
                             TDMI-ST~\cite{feng2023tdmi}  &86.8 &87.4 &85.1 &81.4 &83.8 &82.7 &  78.0 &83.8 \\
                             \hline
                             \rowcolor{mygray} \textbf{VREMD (Ours)} & 87.2 &  \textbf{89.1} & \textbf{85.2} & \textbf{82.4} & \textbf{85.1} & \textbf{83.4} &  \textbf{79.2} &  \textbf{84.5} \\

        \hline
    \end{tabular}
    }
    \caption{Comparisons with the state-of-the-art methods for video pose estimation on the validation sets of the \textbf{PoseTrack2021}~\cite{doering2022posetrack21} dataset.} 
    \label{table:compare_21}
\end{table*}

\begin{figure}[t]
\centering
\includegraphics[width=.99\linewidth]{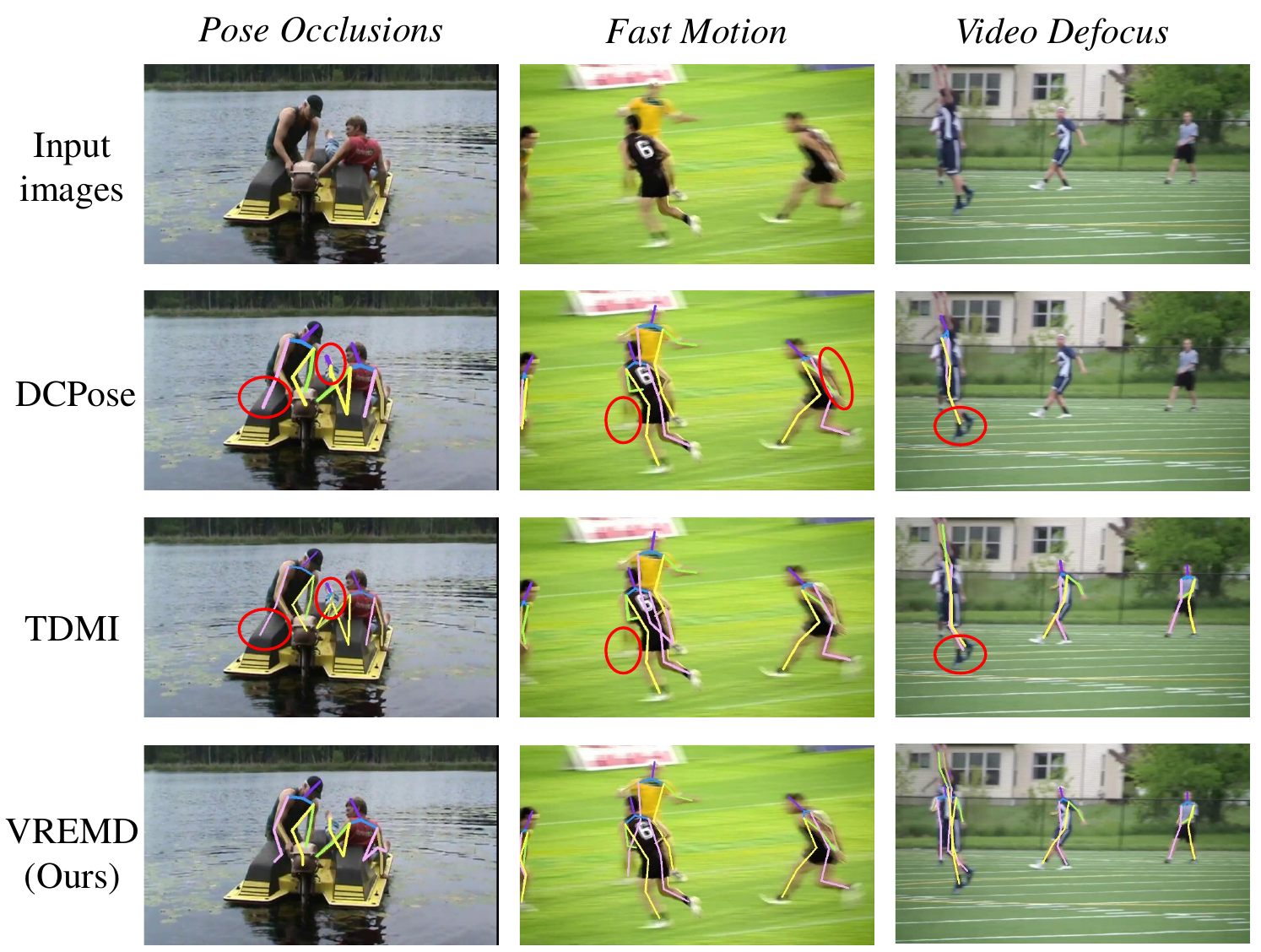}
\caption{Qualitative comparison of our VREMD,  DCPose~\cite{liu2021dcpose}, and TDMI~\cite{feng2023tdmi} on the PoseTrack2017 dataset, featuring challenges such as pose occlusions, fast motion, and video defocus. Red solid circles denote the inaccurate pose predictions.} 
\label{fig:comparison}
\end{figure}

\begin{figure}[htbp]
\centering
\includegraphics[width=.98\linewidth]{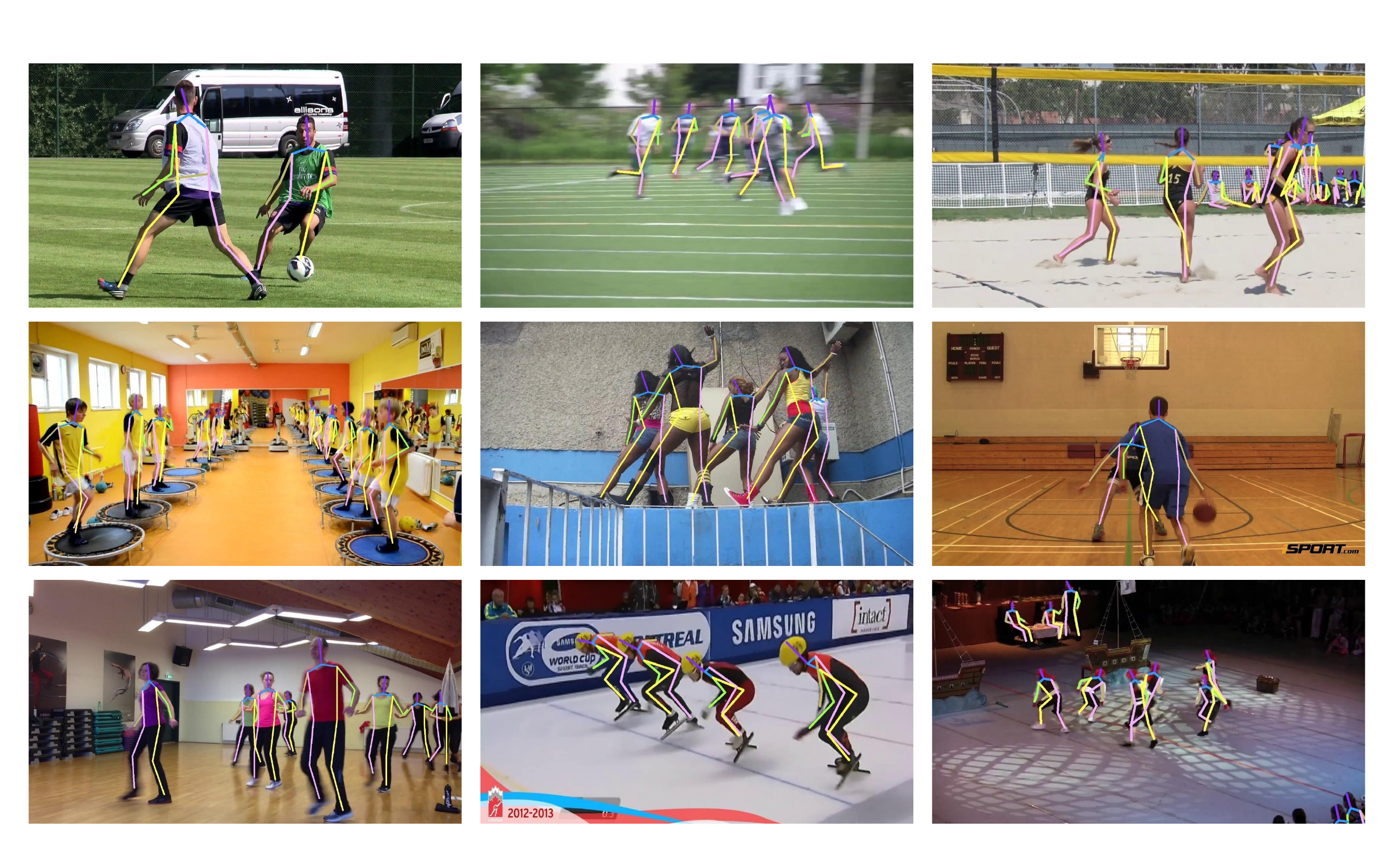}
\caption{Visual results of our VREMD on PoseTrack datasets. Challenging Scenarios such as fast motion or pose occlusion are involved.} 
\label{fig:vis}
\end{figure}

\begin{table}[t] 
    \centering
    \fontsize{9pt}{9pt}\selectfont
         \resizebox{0.8\linewidth}{!}{
        \begin{tabular}{c|cc|c}
            \hline
             Method & HKME  & BMD  & Mean \\
            \hline
            Baseline &  &  &  80.2 \\
            (a) & \checkmark  &    &   85.3   \\
            (b) &   &  \checkmark  &   85.6  \\  
            (c) &  \checkmark &  \checkmark  &  \textbf{87.6}    \\
         
            \hline
        \end{tabular}
      }
    \caption{Ablation of different components in our \textbf{VREMD}.} 
    \label{table:ablation1}
\end{table}

\subsection{Comparison with State-of-the-art Approaches}

\textbf{Results on the PoseTrack2017 Dataset.} We first benchmark our method on the PoseTrack2017~\cite{iqbal2017posetrack} dataset. A total of 22 methods are compared and their performances on the PoseTrack2017 validation set are summarized in Table~\ref{table:compare_17}. Our proposed VREMD consistently outperforms existing state-of-the-art methods, reaching an mAP of 87.6. Compared to the latest top-performing method TDMI-ST~\cite{feng2023tdmi}, our VREMD obtain a 1.7 mAP gain. The performance boost for challenging joints (\textit{i.e.}, wrist, ankle) is also promising: we attain an mAP of 84.8 ($\uparrow$ 3.3) for wrists and an mAP of 81.0 ($\uparrow$ 2.3) for ankles. It is noteworthy that our VREMD operates effectively with fewer input video frames than the most recent works~\cite{liu2022fami,feng2023tdmi}, requiring just three frames as opposed to five. These consistent and substantial improvements in effectiveness indicate the importance of reinforcing the positional attributes of visual representations and integrating joint-related motion dynamics. In addition, we present the visualized results, which include a comparison with existing methods, for scenarios involving complex spatio-temporal interactions (\textit{e.g.}, pose occlusion, blur) in Fig~\ref{fig:comparison}, demonstrating our method's robustness. More visualization results are shown in Figure~\ref{fig:vis}.

\textbf{Results on the PoseTrack2018 Dataset.}  We further evaluate our VREMD on the PoseTrack2018 dataset, and the detailed validation set results are showcased in Table~\ref{table:compare_18}. Once again, as illustrated in this table, our VREMD surpasses all prior state-of-the-art methods, achieving the most exceptional outcomes. We obtain the final performance of 84.6 mAP. The precision for wrists and ankles also shows a noticeable improvement compared to TDMI-ST, scoring 82.1 ($\uparrow$ 1.5) and 79.3 ($\uparrow$ 1.7) respectively.

\textbf{Results on the PoseTrack2021 Dataset.} Performance comparisons of our model and previous state-of-the-art methods on the PoseTrack21 dataset are provided in Table~\ref{table:compare_21}. When evaluated on the PoseTrack2021 validation dataset, the results highlight the outstanding performance of our model. Achieving new state-of-the-art results, our model records an overall mAP of 84.5, outperforming TDMI-ST by a margin of 0.7 mAP. Encouragingly, our method yields a 1.0 mAP improvement over the previous best, attaining 82.4 at the wrist, and shows a 1.2 mAP advance, achieving 79.2 at the ankle, which are recognized as difficult joints to accurately predict. These results, once again, underscore the robustness and superiority of our method in this domain.

\begin{table}[t] 
    \centering
    \fontsize{8pt}{8pt}\selectfont
      \resizebox{0.95\linewidth}{!}{
            \begin{tabular}{c|cc|c}
                \hline
                 Method & Human mask & Keypoint mask & Mean \\
                \hline
                (a) &  &  &  85.9 \\
                (b) & \checkmark  &    &   86.5   \\
                (c) &   &   \checkmark &  86.8  \\
                (d) & \checkmark  & \checkmark  &   \textbf{87.6}    \\
             
                \hline
            \end{tabular}
            }
            \caption{Ablation of various designs in the \textbf{HKME} module.} 
            \label{table:ablation2}
\end{table}

\begin{table}[t] 
    \centering
    \fontsize{8pt}{8pt}\selectfont
        \resizebox{0.99\linewidth}{!}{
        \begin{tabular}{c|cc|cc|c}
                \hline
                 Method & DC & DA & DCA (Ours) & BS (Ours) & Mean \\
                \hline 
                (a) &  \checkmark &  &  & & 84.7 \\
                (b) &  &  \checkmark & &  &    85.8 \\
                (c) &  &  &   \checkmark &   & 87.1   \\
                (d) &   &    &  \checkmark & \checkmark &  \textbf{87.6}   \\
             
                \hline
        \end{tabular}
        }
        \caption{Ablation of various designs in the \textbf{BMD} module.} 
        \label{table:ablation3}
\end{table}

\subsection{Ablation Study}

We carry out extensive ablation studies centered on assessing the impact of individual components within our VREMD architecture, encompassing the Human-Keypoint Mask Enhancement module (HKME) and the Bidirectional Motion Disentanglement module (BMD). We additionally probe into the efficacy of diverse micro-designs incorporated in each module. All experiments are performed on the PoseTrack2017 validation set.

\textbf{Study on components of VREMD.} We experimentally evaluate the effectiveness of each component in our VREMD framework, detailing the quantitative results in Table~\ref{table:ablation1}. Firstly, we establish a baseline for this experiment by coupling a Vision Transformer (ViT) Backbone with a pose detection head. (a) Integrating the Human-Keypoint Mask Enhanced module (HKME) into the baseline yields a substantial gain of 5.1 mAP. This substantial progress indicates that the dual-mask mechanism, offering a coarse-to-fine representation refinement, facilitates improvements in human pose estimation. (b) In the next setup, we exclusively incorporate the Bidirectional Motion Disentanglement module (BMD) into the baseline system. Notably, the Adaptive Deformable Cross (ADC) block, which originally utilized enhanced features from the HKME, now receives backbone output features instead. The outcome achieves an mAP of 85.6, marking an increase of 5.4 mAP. Such a significant boost in performance unequivocally validates the BMD module's proficiency in adaptively excavating bidirectional temporal information, guiding accurate pose estimation. (c) Finally, we incorporate both the HKME and BMD modules into our framework, attaining a culminating performance of 87.6 mAP, which indicates that the synergy of these two components can lead to further enhancements.

\textbf{Study on Human-Keypoint Mask Enhanced module.} We then investigate the impact of the two mask generation techniques in HKME on overall performance. We conduct four experiments and presented them in Table~\ref{table:ablation2}. (a) Generating visual representations using only the spatio-temporal Transformers network. (b) Producing a human mask for coarse filtering of human-related tokens. (c) Calculating a keypoint mask for basic joint token screening. (d) Utilizing dual masks, derived from methods (b) and (c), for the progressive refinement and enhancement of visual tokens, transitioning from coarse to fine detail. This table illustrates that method (a), which does not generate any masks, offers a slight improvement of 0.3 mAP over the setting that removes HKME. Subsequently, applying the human mask alone (b) and the keypoint mask alone (c) achieves respective performances of 86.5 mAP and 86.8 mAP. Although utilizing these masks individually can yield certain accuracy gains, simultaneously employing both for coarse-to-fine representation refinement (d) leads to the optimal results. This promising outcome attests to the superiority of our dual-mask paradigm, which provides a prompt of human joints to the framework, enabling more accurate keypoint localization.

\textbf{Study on Bidirectional Motion Disentanglement module.} Additionally, we explore the influence of our deformable cross attention (DCA) and bidirectional separation strategy. Four experiments are performed and displayed in Table~\ref{table:ablation3}. (a) We first replace our Adaptive Deformable Cross (ADC) block with the deformable conv (DC)~\cite{dai2017deformableconv1}, as adopted in previous works~\cite{liu2021dcpose,liu2022fami,feng2023tdmi}. We observe a slight performance decline, that is, a 0.6 mAP decrease. We speculate that the reason might be the feature map obtained through the attention mechanism is more spatially dispersed and structurally diverse, which is incompatible with the local adaptive variation characteristics of deformable conv. (b) We further employ plain deformable attention (DA)~\cite{zhu2020deformabledetr} and achieve an 85.8 mAP, which proves that deformable attention is more suitable for our frameworks based on attention mechanisms. (c) We propose a novel deformable cross attention (DCA), which integrates the advantages of adaptive receptive field of deformable attention and selective feature highlighting of cross attention, achieving an 87.1 mAP. (d) Finally, we apply a bidirectional separation (BS) strategy to independently capture bidirectional motion dynamics, resulting in a 0.5 mAP improvement, unlike previous methods that concatenate and jointly process bidirectional motion features. These results strongly demonstrate that our method can more effectively capture task-relevant motion cues to facilitate pose estimation.

\section{Related Work}
\textbf{Image-based human pose estimation.} Recent progress in deep learning architectures, as chronicled in ~\cite{dosovitskiy2020imagevit,he2016resnet}, coupled with the proliferation of extensive datasets referenced in ~\cite{lin2014microsoftcoco, shuai2023locate}, has catalyzed the development of a multitude of deep learning methodologies. These methodologies, delineated in ~\cite{sun2019hrnet,wei2016cpm} and proposed for the purpose of image-based human pose estimation, predominantly align with two distinct paradigms: bottom-up and top-down. Bottom-up approaches~\cite{cao2017realtime} initiate with the detection of individual body parts in an image and subsequently attempt to aggregate these parts into a comprehensive human pose. The top-down paradigm~\cite{sun2019hrnet,xu2022vitpose} start by detecting the bounding box around the human body and then localize the target human's keypoints within that area. However, these image-based methods struggle when applied to video streams, since they fail to effectively incorporate the temporal changes between frames. our research builds upon previous image-based approaches, extending them with temporal dynamics capture specifically tailored for video pose estimation.

\textbf{Video-based human pose estimation.} 
 In the early stages, substantial approaches involve utilizing optical flow to establish motion-based assumptions~\cite{pfister2015flowing}. These approaches commonly generate dense optical flow across frames to improve pose heatmap predictions, yet the technique is computationally demanding and prone to errors when faced with marked deterioration in image quality. Recent methods~\cite{he2024dsta} have shifted towards attempting to implicitly capture motion evidence from temporal information by employing deformable convolutions. DCPose~\cite{liu2021dcpose} and PoseWarper~\cite{bertasius2019posewarper} model and process pose temporal residuals and re-refine keypoint detection via multi-scale deformable convolutions for accurate pose estimation. TDMI~\cite{feng2023tdmi} introduces a multi-stage framework that encodes temporal differences for dynamic context modeling, leveraging mutual information to uncover useful temporal clues. Contrary to prior approaches that directly execute feature difference learning in the global space, we strive to enhance visual representations through the aggregation of joint positions, and to dissect representative joint-associated motion dynamics for more robust human pose estimation.

\section{Conclusion and Future Work}

\textbf{Conclusion.} In this paper, we investigate the video-based human pose estimation task from the perspective of local spatial perception and temporal cues disentanglement. A dual-stream architecture is designed to effectively capture spatio-temporal dependencies by collaboratively executing gradual human joint focus and adaptive motion decoupling. Specifically, we present a Human-Keypoint Mask Enhanced module that performs a coarse-to-fine selective representation enhancement to assist the framework in exploring human and joint regions. Additionally, we create a Bidirectional Motion Disentanglement module to adaptively mine pose-related motion evidence. Our method significantly and consistently outperforms state-of-the-art performances on three benchmark datasets: PoseTrack2017, PoseTrack2018, and PoseTrack2021.

\textbf{Limitations and future works.} We identified two limitations in our model: (1) The accuracy of our head joint localization is suboptimal. We believe this is due to good spatial separation of joints but imperfect recognition of their relationships, causing interference from nearby joints, such as the shoulder. We plan to address this by incorporating Graph Neural Networks (GNNs) to better capture these interrelationships. (2) When the target person is severely occluded by others, our method may mistakenly incorporate temporal cues from non-target individuals, reducing pose estimation accuracy. We plan to optimize our visual and motion features using clustering techniques to address this issue.

\section{Acknowledgments} 
This work is supported by the National Natural Science Foundation of China (No. 62372402), and the Key R\&D Program of Zhejiang Province (No. 2023C01217).

\bibliography{aaai25}

\end{document}